\documentclass[journal]{IEEEtran}

\usepackage[utf8]{inputenc}
\usepackage{cite}
\usepackage{amsmath,amssymb,amsfonts}
\usepackage{algorithmic}
\usepackage{algorithm}
\usepackage{textcomp}
\usepackage{xcolor}
\usepackage{booktabs}   
\usepackage{multirow}
\usepackage{url}
\usepackage{siunitx}
\usepackage{caption}
\usepackage{subcaption}

\usepackage{graphicx}

\usepackage[hidelinks]{hyperref}

\newtheorem{proposition}{Proposition}

\begin{document}

\title{Simplex-Optimized Hybrid Ensemble for Large Language Model Text Detection Under Generative Distribution Drift}

\author{Sepyan Purnama Kristanto, Lutfi Hakim, and Dianni Yusuf%
\thanks{The authors are with the Department of Informatics, Politeknik Negeri Banyuwangi, Banyuwangi 68461, Indonesia (e-mail: sepyan@poliwangi.ac.id).}}


\maketitle

\begin{abstract}
The widespread adoption of large language models (LLMs) has made it difficult to distinguish human writing from machine-produced text in many real applications. Detectors that were effective for one generation of models tend to degrade when newer models or modified decoding strategies are introduced. In this work, we study this lack of stability and propose a hybrid ensemble that is explicitly designed to cope with changing generator distributions. The ensemble combines three complementary components: a RoBERTa-based classifier fine-tuned for supervised detection, a curvature-inspired score based on perturbing the input and measuring changes in model likelihood, and a compact stylometric model built on hand-crafted linguistic features. The outputs of these components are fused on the probability simplex, and the weights are chosen via validation-based search. We frame this approach in terms of variance reduction and risk under mixtures of generators, and show that the simplex constraint provides a simple way to trade off the strengths and weaknesses of each branch. Experiments on a 30\,000-document corpus drawn from several LLM families including models unseen during training and paraphrased attack variants show that the proposed method achieves 94.2\% accuracy and an AUC of 0.978. The ensemble also lowers false positives on scientific articles compared to strong baselines, which is critical in educational and research settings where wrongly flagging human work is costly.
\end{abstract}


\section{Introduction}
Text generated by large language models (LLMs) is now routinely used in homework, reports, programming, and informal communication. Many of these uses are benign, but the same technology can undermine assessment, authorship, or review processes when it is employed without disclosure. As a result, there is growing interest in tools that can provide a probability that a given text was written with the assistance of an LLM.

At first glance, this problem appears similar to other binary classification tasks. In practice, however, one key assumption often fails: the distribution of the positive class, namely machine-generated text, changes quickly over time. When a detector is trained on text from one system and later evaluated on text from newer models with different training objectives or decoding strategies, the performance often drops sharply. For example, a detector tuned on GPT-3.5 text may not behave well when facing GPT-4 or Claude outputs, especially after paraphrasing or editing.

In standard supervised learning, it is common to assume that both training and test data are drawn from the same distribution. In the detection setting, the human class $Y=0$ is relatively stable, but the machine class $Y=1$ can be viewed as a family of distributions indexed by model architecture, prompt formulation, and decoding parameters. We refer to this situation as \emph{generative distribution drift}. It is a form of covariate shift where the change is driven by external model updates rather than by a natural temporal process.

Several lines of work have been proposed to handle this task. Fine-tuned transformer classifiers treat the problem as a standard supervised task and can achieve high accuracy when training and test generators are aligned. Methods based on model likelihoods and token-level statistics attempt to avoid retraining by looking at how probable a text is under a reference model. Stylometric methods focus on surface features and writing patterns that differ between human and machine text. Each of these families has clear advantages but also breaks down under certain conditions, for instance when text is short, heavily edited, or produced by an unseen model.

In this paper, we take a pragmatic view and ask whether a simple ensemble of heterogeneous detectors can improve robustness to generator drift without introducing an overly complex meta-architecture. Our starting point is that the different components tend to make different kinds of errors. We therefore combine them in a convex mixture on the probability simplex and tune the weights to balance their contributions.

The main contributions of this work are:
\begin{itemize}
    \item We formalize LLM text detection as classification under shifting generator distributions and discuss how this affects risk and variance for detectors deployed over time.
    \item We introduce a hybrid ensemble that combines a supervised RoBERTa model, a perturbation-based curvature score, and a stylometric classifier, with fusion weights restricted to the probability simplex and selected by grid search.
    \item We build a 30\,000-document dataset (GenDrift-30K) that separates in-distribution and out-of-distribution generators, including paraphrased variants, and use it to systematically evaluate cross-generator generalization and false positive rates on academic text.
    \item We present empirical evidence that the simplex-constrained ensemble provides consistent gains over individual components and naïve averaging, particularly under drift scenarios and adversarial paraphrasing.
\end{itemize}

The rest of the paper is structured as follows. Section~II reviews related work on detection and ensemble methods. Section~III presents the theoretical framing. Section~IV describes the architecture and components of the ensemble. Section~V outlines the dataset and experimental protocol. Section~VI reports results, and Section~VII discusses implications and limitations. Section~VIII concludes the paper.

\section{Related Work}

\subsection{Supervised Detection with Pretrained Transformers}
Early work on neural text detection focused on fine-tuning pretrained encoders such as BERT and RoBERTa on labeled corpora of human and machine text. Solaiman \textit{et al.} \cite{solaiman2019release} reported that a RoBERTa model trained on GPT-2 outputs could distinguish generated text with high accuracy when evaluated on similar data. Subsequent studies extended this idea to newer generators and more varied domains.

However, the generalization properties of such detectors are fragile. Chakraborty \textit{et al.} \cite{chakraborty2023possibilities} and Sadasivan \textit{et al.} \cite{sadasivan2023} observed that performance can drop to near-chance when the detector is evaluated on text from model families not seen during training. Guo \textit{et al.} \cite{guo2023close} analyzed this phenomenon in more detail and showed that detectors often learn correlations tied to specific token distributions or stylistic quirks of the training generator. When these cues disappear in newer models, the learned decision boundary becomes less meaningful.

\subsection{Likelihood-Based and Zero-Shot Approaches}
As an alternative to supervised training, several methods aim to detect machine text by inspecting likelihood statistics from a reference language model. GLTR \cite{gehrmann2019gltr} visualizes token rank distributions to help humans spot unusual patterns. Mitchell \textit{et al.} \cite{mitchell2023detectgpt} proposed DetectGPT, which perturbs the input text and uses changes in log-likelihood to estimate a curvature score. The underlying intuition is that generated text tends to lie near regions of locally high likelihood compared to perturbed variants.

These approaches can be applied without retraining and may transfer across tasks, but they have limitations. Access to token probabilities or logits is not always available, especially for proprietary LLMs. Moreover, their effectiveness depends on the reference model and the length and structure of the input. Krishna \textit{et al.} \cite{krishna2023paraphrasing} showed that paraphrasing operations can substantially reduce the effectiveness of several detection schemes, illustrating that likelihood-based cues can be altered without changing the underlying content.

\subsection{Stylometric and Feature-Based Methods}
Stylometric analysis predates neural language models and has long been used for authorship attribution and forensic linguistics. Classic features include type token ratios, sentence length statistics, function word frequencies, and syntactic patterns. Recent work has applied similar ideas to LLM detection. Frankenstein and Grimmer \cite{frankenstein2023stylometric} studied stylometric markers for AI-generated text in realistic settings with heterogeneous generators.

Feature-based methods are attractive because they are transparent and computationally efficient. They can also be combined with simple classifiers such as support vector machines or random forests. On the other hand, their capacity is limited compared to large neural encoders, and their performance can lag on domains where surface characteristics are similar across human and machine text.

\subsection{Ensembles and Robustness in NLP}
Ensemble techniques are a standard tool for improving predictive performance and stability. Dietterich \cite{dietterich2000} and Kuncheva \cite{kuncheva2004} analyzed conditions under which combining models is beneficial, emphasizing that base learners should be both reasonably accurate and diverse. In the context of deep learning, Lakshminarayanan \textit{et al.} \cite{lakshminarayanan2017simple} used deep ensembles to improve uncertainty estimates. For out-of-distribution detection, Hendrycks and Gimpel \cite{hendrycks2017baseline} explored simple confidence-based methods that can be combined with ensembles.

In NLP, ensembles are often implemented as either majority votes over several models or as stacking, where a meta-classifier learns how to aggregate base predictions \cite{wolpert1992stacked}. While flexible, stacking can itself be prone to overfitting. In this work, we follow a simpler strategy: we restrict fusion weights to the probability simplex and search over this low-dimensional space using a validation set. This gives a lightweight way to balance the contributions of several heterogeneous components.

\section{Theoretical Framework}

\subsection{Problem Setting and Notation}
Let $\mathcal{X}$ denote the space of text documents, and let $Y \in \{0,1\}$ denote a binary label, where $Y=0$ corresponds to human-written text and $Y=1$ to machine-generated text. We assume that the human distribution $P_H(X) = P(X \mid Y=0)$ is relatively stable over the time scale of interest, for example several semesters in a university.

Machine-generated text, in contrast, is not represented by a single distribution. We write $P_\theta(X) = P(X \mid Y=1, \theta)$ for the distribution of text produced by a generator indexed by $\theta$, where $\theta$ encodes architecture, training, prompting, and decoding configuration. The set of possible generators is denoted $\Theta$, and we write
\begin{equation}
\mathcal{P}_M = \{P_\theta : \theta \in \Theta\}.
\end{equation}
A deployed detector will in practice encounter a mixture of these machine distributions, but the mixing weights can change as new systems become available.

\subsection{Generative Distribution Drift}
Suppose a detector is trained on samples from a fixed generator $\theta_{\mathrm{src}}$. The source machine distribution is then $P_{\theta_{\mathrm{src}}}$. At deployment time, however, the generator index may be $\theta_{\mathrm{tgt}}$, which can differ significantly. The training and target joint distributions can be written as
\begin{equation}
P_{\mathrm{src}}(X,Y), \quad P_{\mathrm{tgt}}(X,Y),
\end{equation}
with $P_{\mathrm{src}}(X \mid Y=1) = P_{\theta_{\mathrm{src}}}(X)$ and $P_{\mathrm{tgt}}(X \mid Y=1) = P_{\theta_{\mathrm{tgt}}}(X)$.

If a classifier $f:\mathcal{X} \rightarrow [0,1]$ is trained to minimize empirical risk on the source distribution, its target error can be bounded using standard domain adaptation arguments:
\begin{equation}
\epsilon_{\mathrm{tgt}}(f) \leq \epsilon_{\mathrm{src}}(f) + \frac{1}{2} d_{\mathcal{H}\Delta\mathcal{H}}(P_{\mathrm{src}}, P_{\mathrm{tgt}}) + \lambda,
\end{equation}
where $d_{\mathcal{H}\Delta\mathcal{H}}$ is the $\mathcal{H}\Delta\mathcal{H}$-divergence and $\lambda$ is the minimal combined risk over hypothesis class $\mathcal{H}$. When the machine distributions differ substantially, the divergence term can be large, and even a well-trained detector on the source may perform poorly on the target.

In our context, we are not in a position to control the evolution of $P_\theta$. Instead, we seek detector families and aggregation strategies that depend on features which are less sensitive to particular values of $\theta$, and that can provide good performance across a range of generators.

\subsection{Ensemble Variance and Covariance}
Consider $M$ component detectors $f_1,\dots,f_M$, each mapping $\mathcal{X}$ to $[0,1]$. We combine them in a convex mixture
\begin{equation}
\bar{f}(x) = \sum_{m=1}^M w_m f_m(x),
\end{equation}
where $\mathbf{w} = (w_1,\dots,w_M)$ lies on the simplex
\begin{equation}
\Delta_M = \left\{ \mathbf{w} \in \mathbb{R}^M : w_m \geq 0,\ \sum_{m=1}^M w_m = 1 \right\}.
\end{equation}
Under a fixed data distribution, the variance of the ensemble predictions can be expressed as
\begin{equation}
\mathrm{Var}(\bar{f}) = \sum_{m=1}^M w_m^2 \mathrm{Var}(f_m) + \sum_{m \neq n} w_m w_n \mathrm{Cov}(f_m, f_n).
\end{equation}

\begin{proposition}
If the component detectors $f_m$ are not perfectly correlated and rely on different types of information (for instance, semantic, probabilistic, and stylometric cues), then there exist weights $\mathbf{w} \in \Delta_M$ such that $\mathrm{Var}(\bar{f})$ is strictly smaller than the variance of each component.
\end{proposition}

The proof follows standard ensemble theory: as long as each component is better than random and their errors are not fully aligned, averaging reduces variance. The key practical question is how to choose $\mathbf{w}$ so that we also control the bias and performance under distribution shift.

\subsection{Risk Under Generator Mixtures}
At deployment, a detector observes machine text drawn not from a single $P_\theta$ but from a mixture
\begin{equation}
P_M^{(\pi)}(x) = \int_{\Theta} P_\theta(x)\,d\pi(\theta),
\end{equation}
where $\pi$ is a mixing distribution over generators. The mixing distribution itself may be unknown and may evolve over time. For a detector $f$, we define the risk under a mixture $\pi$ as
\begin{equation}
\mathcal{R}(f;\pi) = \mathbb{E}_{x \sim P_H}[\ell(f(x),0)] + \mathbb{E}_{x \sim P_M^{(\pi)}}[\ell(f(x),1)],
\end{equation}
with $\ell$ denoting binary cross-entropy loss. Since $\pi$ is not observed, we are interested in the worst-case risk over a plausible set $\Pi$ of mixing distributions:
\begin{equation}
\mathcal{R}^{\mathrm{wc}}(f) = \sup_{\pi \in \Pi} \mathcal{R}(f;\pi).
\end{equation}

\begin{proposition}
Consider an ensemble $\bar{f}_{\mathbf{w}} = \sum_m w_m f_m$ with $\mathbf{w} \in \Delta_M$. Assume that for each generator index $\theta$, there exists at least one component $f_{m(\theta)}$ whose risk on $P_\theta$ is not worse than the other components. If the variation of component risks across $\theta$ is not perfectly aligned, then there exists a weight vector $\mathbf{w}^\star \in \Delta_M$ such that
\begin{equation}
\mathcal{R}^{\mathrm{wc}}(\bar{f}_{\mathbf{w}^\star}) \leq \min_m \mathcal{R}^{\mathrm{wc}}(f_m).
\end{equation}
\end{proposition}

A sketch of the argument is as follows. For each $\theta$, the ensemble risk is a convex combination of component risks. Under the stated assumptions, the worst-case mixture over $\Pi$ is bounded by the best convex combination of these risks, and the minimum over the simplex cannot exceed the minimum worst-case risk over individual components. In practice, we approximate $\mathbf{w}^\star$ by searching over a coarse grid of the simplex using a validation set that approximates the diversity of generators.

This theoretical framing motivates the concrete design in the next section: instead of relying on a single strong but brittle detector, we combine three different mechanisms and select simplex weights using held-out data that include both in-distribution and out-of-distribution examples.

\section{Methodology}

\subsection{System Overview}
The proposed system consists of three parallel analysis branches followed by a fusion module. An input document is first normalized and then passed to:
\begin{enumerate}
    \item a RoBERTa-based classifier that captures semantic and contextual information (M1),
    \item a perturbation-based curvature estimator derived from log-likelihood differences (M2), and
    \item a stylometric feature extractor and random forest classifier (M3).
\end{enumerate}
Each branch outputs a probability that the text is machine-generated. The fusion module then forms a weighted sum of these probabilities, with weights restricted to $\Delta_3$ and chosen through grid search on a validation set.

\begin{figure}[t!]
\centering
\includegraphics[width=\columnwidth, height=3cm]{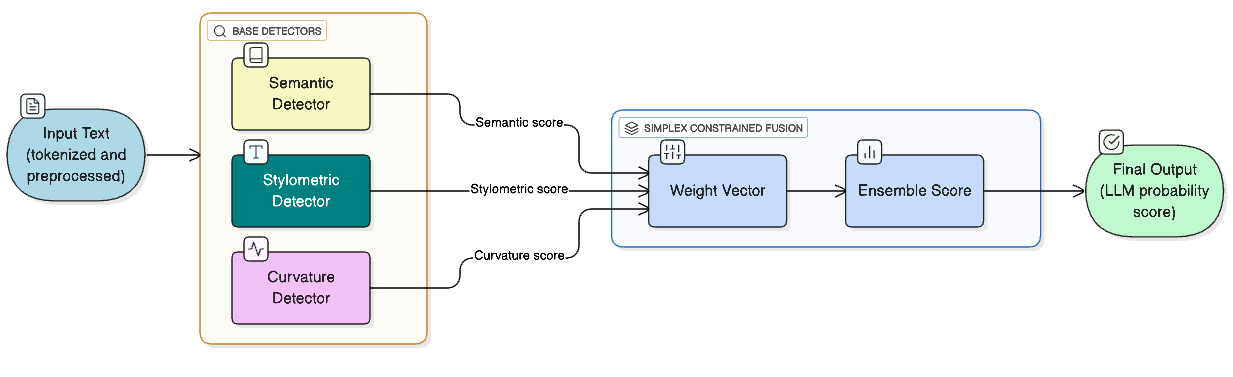}
\caption{Overview of the proposed Simplex-Optimized Hybrid Ensemble. The document is processed by three detectors (semantic, curvature-based, and stylometric). Their outputs are combined using weights constrained to the probability simplex and tuned on a validation set.}
\label{fig:architecture}
\end{figure}

Fig.~\ref{fig:architecture} provides a schematic view of the pipeline.

\subsection{Semantic Detector (M1: RoBERTa)}
The first component is a RoBERTa-base encoder fine-tuned for binary classification. The input text is tokenized and truncated or padded to a fixed maximum length. RoBERTa processes the sequence and produces contextual embeddings, including a special classification token representation $h_{\texttt{[CLS]}}$. We compute
\begin{equation}
p_1(x) = \sigma\big(\mathbf{w}_{\mathrm{cls}}^\top \mathrm{Dropout}(h_{\texttt{[CLS]}}) + b_{\mathrm{cls}}\big),
\end{equation}
where $\sigma$ is the logistic function. The model is fine-tuned with binary cross-entropy loss on a training set containing both human and machine text.

RoBERTa is effective at capturing longer-range coherence and semantic patterns. However, as noted in prior work, its performance can deteriorate when the generator or domain changes significantly. For this reason, it is treated as one component rather than the sole detector.

\subsection{Curvature-Based Detector (M2)}
The second component approximates the idea introduced by DetectGPT \cite{mitchell2023detectgpt}. Given a reference language model with scoring function $q(\cdot)$ (e.g., log-likelihood under GPT-2), we consider the original text $x$ and a set of $k$ perturbed variants $\{\tilde{x}_i\}_{i=1}^k$. The perturbations are produced by masking or editing spans using a text-to-text model such as T5.

We compute a standardized curvature statistic
\begin{equation}
S(x) = \frac{q(x) - \frac{1}{k} \sum_{i=1}^k q(\tilde{x}_i)}{\sqrt{\mathrm{Var}\big(q(\tilde{x}_i)\big)}}.
\end{equation}
Intuitively, if $x$ lies near a local optimum of the reference model's likelihood surface, then $q(x)$ will tend to be higher than the scores of its perturbations, leading to a large positive $S(x)$. This behavior is more common for machine-generated text.

To integrate this score with other components, we map $S(x)$ to a probability $p_2(x)$ using a logistic calibration model
\begin{equation}
p_2(x) = \sigma(a S(x) + c),
\end{equation}
with parameters $(a,c)$ learned on the validation set via Platt scaling \cite{platt1999probabilistic}.

\subsection{Stylometric Detector (M3)}
The third branch focuses on simple, interpretable stylistic features. For each document, we compute a feature vector $\mathbf{v} \in \mathbb{R}^5$ consisting of:
\begin{itemize}
    \item \emph{Type Token Ratio (TTR)}: the ratio of distinct word types to total tokens, capturing vocabulary variety.
    \item \emph{Burstiness}: the variance of sentence lengths measured in tokens.
    \item \emph{Perplexity}: estimated under a baseline $n$-gram model trained on human text.
    \item \emph{Stopword Divergence}: the Kullback Leibler divergence between the document's stopword distribution and a human reference distribution.
    \item \emph{Syntactic Depth}: the average depth of dependency parse trees computed over sentences.
\end{itemize}

These features are fed into a random forest classifier with 200 trees and class-balanced weights. The classifier outputs a probability $p_3(x)$ that the text is machine-generated. While this model is weaker than RoBERTa on in-distribution data, it behaves differently on scientific text and paraphrased outputs, making it a useful complement in the ensemble.

\subsection{Simplex-Constrained Fusion}
Given the three component probabilities
\begin{equation}
\mathbf{p}(x) = [p_1(x), p_2(x), p_3(x)]^\top,
\end{equation}
the ensemble prediction is defined as
\begin{equation}
\hat{y}_{\mathrm{ens}}(x) = \mathbf{w}^\top \mathbf{p}(x),
\end{equation}
with $\mathbf{w} \in \Delta_3$. We choose $\mathbf{w}$ by maximizing the F1-score on a validation set:
\begin{equation}
\mathbf{w}^\star = \arg\max_{\mathbf{w} \in \Delta_3} \mathrm{F1}(\mathbf{w}).
\end{equation}

Because the dimension is small, we simply perform a grid search over the simplex using a fixed step size. Specifically, we let $w_1$ and $w_2$ vary in $\{0, 0.05, 0.10, \dots, 1\}$ under the constraint $w_1 + w_2 \leq 1$, and set $w_3 = 1 - w_1 - w_2$. For each candidate $\mathbf{w}$, we compute the ensemble predictions on the validation set, threshold at 0.5, and record the resulting F1-score. The best-performing weight vector is retained.

Algorithm~\ref{alg:simplex} summarizes the search procedure.

\begin{algorithm}[t!]
\caption{Grid Search for Simplex Weights}
\label{alg:simplex}
\begin{algorithmic}[1]
\STATE \textbf{Input:} Validation probabilities $\mathbf{P}_{\mathrm{val}} \in \mathbb{R}^{N \times 3}$, labels $\mathbf{Y}_{\mathrm{val}}$
\STATE \textbf{Output:} Optimal weights $\mathbf{w}^\star$
\STATE $F1_{\mathrm{best}} \gets 0$, $\mathbf{w}^\star \gets [1/3, 1/3, 1/3]^\top$
\FOR{$w_1$ in $\{0, 0.05, \dots, 1\}$}
    \FOR{$w_2$ in $\{0, 0.05, \dots, 1-w_1\}$}
        \STATE $w_3 \gets 1 - w_1 - w_2$
        \STATE $\mathbf{w} \gets [w_1, w_2, w_3]^\top$
        \STATE $\hat{\mathbf{y}} \gets \mathbb{I}(\mathbf{P}_{\mathrm{val}} \mathbf{w} > 0.5)$
        \STATE $F1 \gets \text{F1-Score}(\hat{\mathbf{y}}, \mathbf{Y}_{\mathrm{val}})$
        \IF{$F1 > F1_{\mathrm{best}}$}
            \STATE $F1_{\mathrm{best}} \gets F1$
            \STATE $\mathbf{w}^\star \gets \mathbf{w}$
        \ENDIF
    \ENDFOR
\ENDFOR
\RETURN $\mathbf{w}^\star$
\end{algorithmic}
\end{algorithm}

In our experiments, the optimal weights are typically close to $[0.55, 0.30, 0.15]^\top$, indicating that the semantic detector carries the largest weight but the other two branches still play a substantial role, especially under drift.

\section{Experimental Setup}

\subsection{GenDrift-30K Dataset}
To assess performance under distribution drift, we constructed a dataset that intentionally separates training, validation, and test generators. The resulting corpus, which we call GenDrift-30K, contains 30\,000 documents.

Table~\ref{tab:dataset} summarizes the composition.

\begin{table*}[t!]
\centering
\caption{Composition of the GenDrift-30K Dataset. Training uses only GPT-3.5 and LLaMA-2 as machine sources. The test split includes unseen generators and paraphrased attacks to probe robustness.}
\label{tab:dataset}
\renewcommand{\arraystretch}{1.2}
\begin{tabular}{lllccl}
\toprule
\textbf{Split} & \textbf{Class} & \textbf{Source / Generator} & \textbf{Count} & \textbf{Avg. Length} & \textbf{Description} \\
\midrule
\multirow{2}{*}{Train} 
 & Human & Reuters, Wikipedia, arXiv & 5\,000 & 520 words & Mixed news, encyclopedic entries, and scientific abstracts. \\
 & Machine & GPT-3.5 Turbo, LLaMA-2 7B & 5\,000 & 515 words & Instruction-following text and expository outputs. \\
\midrule
\multirow{2}{*}{Validation} 
 & Human & Student essays (local) & 2\,500 & 480 words & Academic-style essays with varying proficiency. \\
 & Machine & GPT-3.5, Mistral 7B & 2\,500 & 490 words & Outputs from seen and partially unseen models. \\
\midrule
\multirow{4}{*}{Test (Drift)} 
 & Human & arXiv (unseen papers) & 5\,000 & 600 words & Full paragraphs from scientific articles. \\
 & Machine A & GPT-4 & 3\,000 & 610 words & Drift Scenario 1: newer proprietary model. \\
 & Machine B & Claude 3 Opus & 1\,000 & 580 words & Drift Scenario 2: alternative architecture. \\
 & Machine C & PEGASUS (paraphrased) & 1\,000 & 550 words & Drift Scenario 3: paraphrased attacks on base outputs. \\
\bottomrule
\end{tabular}
\end{table*}

The training split is limited to text generated by GPT-3.5 Turbo and LLaMA-2, along with human documents drawn from public corpora. The validation split contains machine text from GPT-3.5 and Mistral 7B, and human texts from student essays, which better reflect academic assignments. The test split is designed to represent drift: it includes GPT-4 and Claude outputs, human scientific text from arXiv that was not used during training, and machine text that has been paraphrased using PEGASUS \cite{zhang2020pegasus}.

\subsection{Prompting and Filtering Procedures}
For machine-generated examples, we used a set of prompt templates covering argumentative essays, technical explanations, short answers, and narrative descriptions. Prompts were varied to avoid repeated phrasings. We manually inspected a subset of the outputs and applied simple rules to remove:
\begin{itemize}
    \item responses that explicitly state that they are produced by an AI assistant;
    \item outputs that contain boilerplate safety or policy disclaimers;
    \item texts that are excessively short (less than 150 words) or dominated by code or lists.
\end{itemize}

Human texts from news sources and Wikipedia were sampled at the paragraph level, with care taken to avoid fragments that are mostly tabular or purely numeric. For student essays, we anonymized the data and removed personal information. We used plagiarism checking tools and web search to filter out essays that appear to be heavily copied from online sources.

For arXiv documents, we extracted contiguous paragraphs from introductions and related work sections of papers in computer science and related fields. We excluded sections that are primarily equations or references to avoid trivial cues.

\subsection{Implementation Details}
All neural models were implemented using the Hugging Face Transformers library. The main training and evaluation settings are as follows:
\begin{itemize}
    \item \textbf{RoBERTa (M1):} RoBERTa-base, maximum sequence length of 512 tokens, learning rate $2 \times 10^{-5}$, batch size 16, three epochs, Adam optimizer with linear warm-up, and early stopping based on validation AUC.
    \item \textbf{Curvature (M2):} Reference model GPT-2 Medium for scoring, $k=20$ perturbed variants per document, span length of two tokens for masking, and logistic calibration fitted on validation data.
    \item \textbf{Stylometry (M3):} Random forest with 200 trees, maximum depth 20, Gini impurity criterion, and class-balanced sampling.
    \item \textbf{Hardware:} Training for RoBERTa and computing curvature scores were carried out on a single NVIDIA A100 GPU, while stylometric feature extraction and random forest training were run on a CPU machine.
\end{itemize}

We report average metrics over multiple random seeds for the training procedure to reduce noise, although variation across runs was small.

\section{Results and Analysis}

\subsection{Overall Performance}
We first compare the proposed ensemble to its individual components and to a simple TF-IDF + SVM baseline. Table~\ref{tab:main} reports accuracy, precision, recall, F1-score, and AUC on the combined test set that includes all human and machine examples from the drift split.

\begin{table}[t!]
\centering
\caption{Performance on the Combined Drift Test Set. The proposed Simplex Hybrid ensemble consistently outperforms individual detectors and a naïve averaging scheme.}
\label{tab:main}
\renewcommand{\arraystretch}{1.2}
\begin{tabular}{lccccc}
\toprule
\textbf{Model} & \textbf{Acc.} & \textbf{Prec.} & \textbf{Rec.} & \textbf{F1} & \textbf{AUC} \\
\midrule
TF-IDF + SVM         & 75.4 & 72.1 & 78.5 & 0.751 & 0.831 \\
Stylometric (M3)      & 78.5 & 81.2 & 65.4 & 0.780 & 0.850 \\
Curvature (M2)        & 87.1 & 91.5 & 82.3 & 0.869 & 0.925 \\
RoBERTa (M1)          & 89.2 & 88.4 & 90.1 & 0.890 & 0.945 \\
Average Ensemble      & 92.5 & 93.1 & 91.8 & 0.924 & 0.967 \\
\textbf{Simplex Hybrid} & \textbf{94.2} & \textbf{95.4} & \textbf{93.1} & \textbf{0.941} & \textbf{0.978} \\
\bottomrule
\end{tabular}
\end{table}

The Simplex Hybrid ensemble improves accuracy by about 1.7 percentage points over the naive average of component probabilities and by 5.0 percentage points over the best single component (RoBERTa). The gain in AUC is particularly notable, reflecting more consistent performance across decision thresholds.

\subsection{Cross-Generator Generalization}
To better understand behavior under distribution drift, we break down performance by generator on the machine portion of the test set. Table~\ref{tab:drift} reports accuracy for RoBERTa, the curvature-based detector, and the ensemble, along with the gain of the ensemble over the best single detector for each generator.

\begin{table}[t!]
\centering
\caption{Accuracy for Different Machine Generators. The ensemble maintains higher accuracy than individual components, especially on paraphrased attacks.}
\label{tab:drift}
\renewcommand{\arraystretch}{1.2}
\begin{tabular}{lccc|c}
\toprule
\textbf{Generator} & \textbf{RoBERTa} & \textbf{Curvature} & \textbf{Hybrid} & \textbf{Gain} \\
\midrule
\textit{In-distribution} &&&& \\
GPT-3.5 Turbo & 90.1\% & 88.2\% & \textbf{95.3\%} & +5.2\% \\
\midrule
\textit{Out-of-distribution} &&&& \\
GPT-4          & 87.5\% & 85.4\% & \textbf{91.6\%} & +4.1\% \\
Claude 3 Opus  & 86.8\% & 84.1\% & \textbf{90.9\%} & +4.1\% \\
PEGASUS (attack) & 73.1\% & 75.5\% & \textbf{87.3\%} & \textbf{+14.2\%} \\
\bottomrule
\end{tabular}
\end{table}

The results show that the ensemble consistently outperforms its components across generators. The largest relative improvement is observed for the paraphrased attack setting, where the single detectors experience substantial drops. The stylometric and curvature-based components appear to provide complementary information that helps the ensemble resist these attacks.

\subsection{False Positives on Academic Text}
For many users, a key concern is the rate at which detectors erroneously label human-written academic text as machine-generated. We therefore measure the false positive rate (FPR) on the subset of 1\,000 human documents drawn from arXiv.

Table~\ref{tab:fpr} compares the FPR for several detectors.

\begin{table}[t!]
\centering
\caption{False Positive Rate (FPR) on Human Scientific Articles. Lower values are better.}
\label{tab:fpr}
\renewcommand{\arraystretch}{1.2}
\begin{tabular}{lcc}
\toprule
\textbf{Model} & \textbf{FPR (\%)} & \textbf{Relative Change} \\
\midrule
TF-IDF + SVM       & 18.5 & -- \\
RoBERTa (M1)        & 8.9  & baseline \\
Curvature (M2)      & 9.8  & +10\% \\
\textbf{Hybrid Ensemble} & \textbf{5.8}  & \textbf{--35\% vs.\ RoBERTa} \\
\bottomrule
\end{tabular}
\end{table}

The ensemble achieves a substantially lower FPR than RoBERTa alone. Manual inspection indicates that RoBERTa tends to over-flag texts with highly structured, low-perplexity phrasing, which often occurs in scientific writing. The stylometric branch, which explicitly models structural complexity, helps correct some of these cases.

\subsection{ROC Curves and Calibration}
\begin{figure}[t!]
    \centering
    \begin{subfigure}[b]{0.48\columnwidth}
        \includegraphics[width=\linewidth]{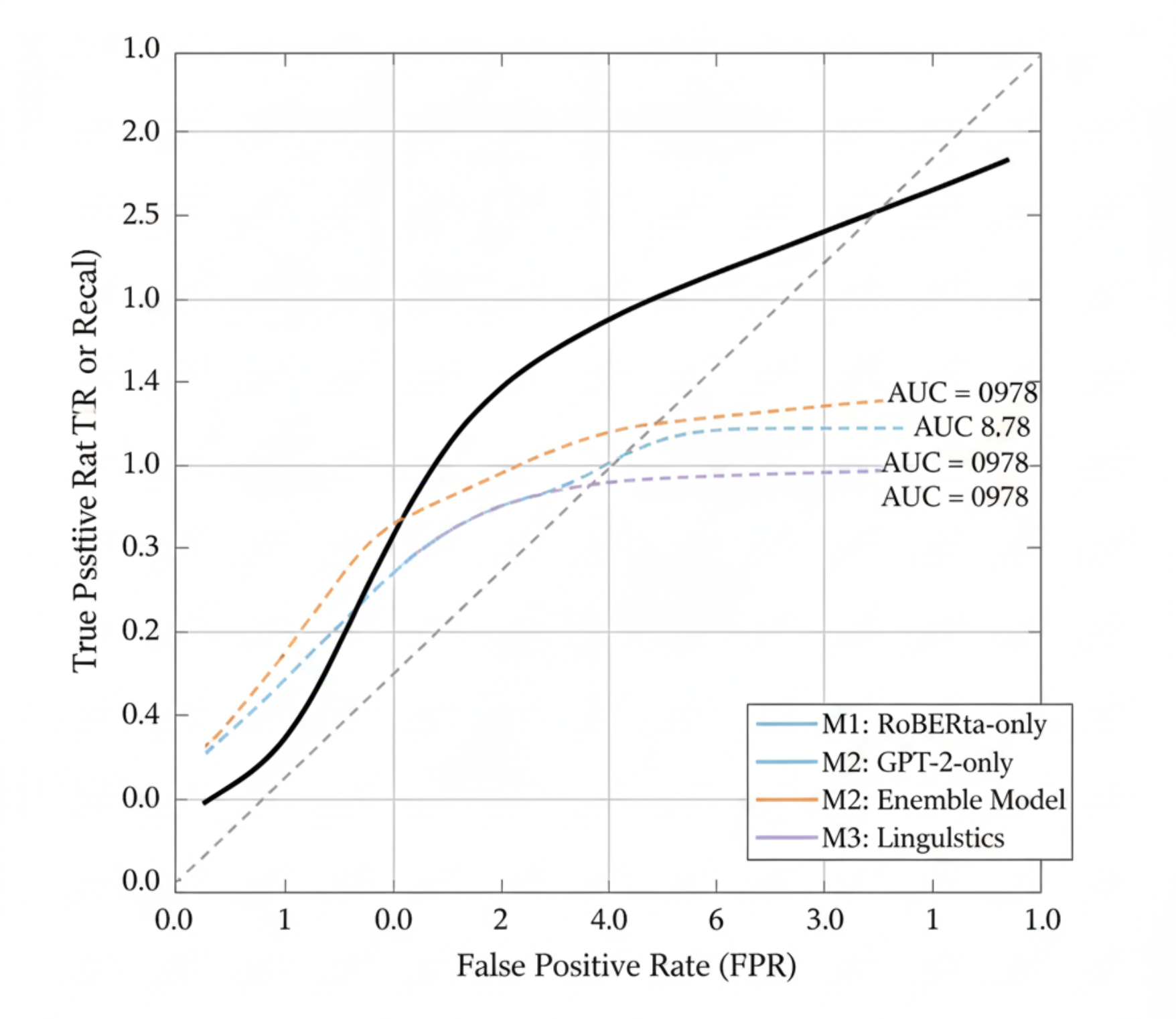}
        \caption{ROC curve}
        \label{fig:roc}
    \end{subfigure}
    \hfill
    \begin{subfigure}[b]{0.48\columnwidth}
        \includegraphics[width=\linewidth]{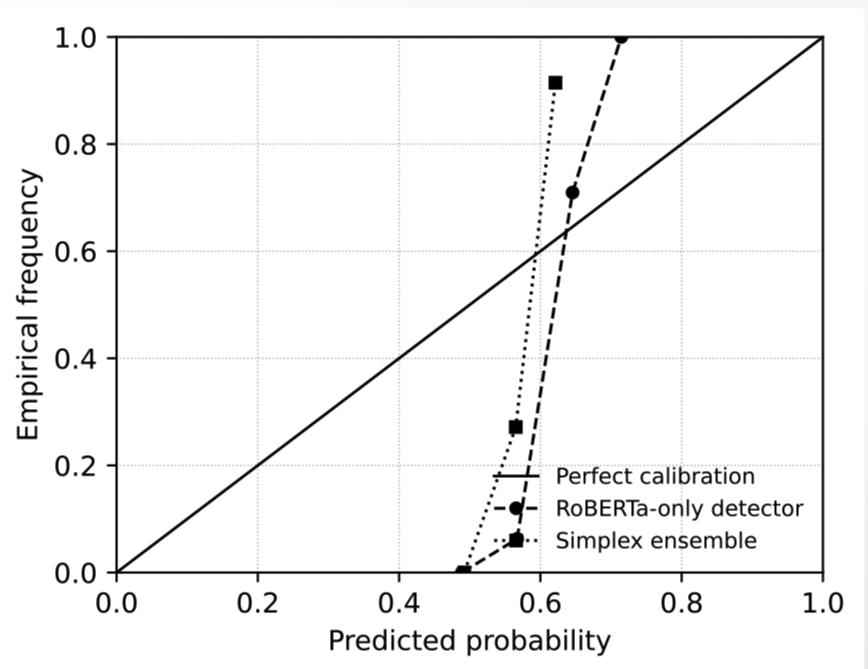}
        \caption{Calibration plot}
        \label{fig:calib}
    \end{subfigure}
    \caption{(a) ROC curves for single detectors and the ensemble. The ensemble dominates across thresholds. (b) Reliability diagram on the test set, showing that the ensemble produces better-calibrated probabilities than the RoBERTa-only detector.}
    \label{fig:plots}
\end{figure}

To further assess the quality of the probability outputs, we examine ROC and calibration curves. Fig.~\ref{fig:plots} shows the ROC curve for each detector and the ensemble, as well as a reliability diagram comparing predicted probabilities to empirical frequencies.

The ROC curve confirms the AUC improvement reported in Table~\ref{tab:main}. The calibration plot shows that RoBERTa tends to produce overconfident predictions, whereas the ensemble probabilities are closer to the diagonal, indicating better alignment between scores and observed frequencies. For use as a screening tool rather than a final decision-maker, well-calibrated probabilities can be more informative than hard labels \cite{niculescu2005predicting}.

\subsection{Ablation Study}
To understand the contribution of each branch, we conduct an ablation study where we remove one component at a time from the ensemble and renormalize the remaining weights. Table~\ref{tab:ablation} shows accuracy, F1, AUC, and academic FPR when each component is omitted.

\begin{table}[t!]
\centering
\caption{Ablation Study on the Drift Test Set. Removing any branch leads to a drop in performance, confirming that the three components provide complementary signals.}
\label{tab:ablation}
\renewcommand{\arraystretch}{1.2}
\begin{tabular}{lcccc}
\toprule
\textbf{Configuration} & \textbf{Acc.} & \textbf{F1} & \textbf{AUC} & \textbf{FPR\_Acad.} \\
\midrule
Full (M1+M2+M3)       & \textbf{94.2} & \textbf{0.941} & \textbf{0.978} & \textbf{5.8\%} \\
Without M1 (RoBERTa)  & 89.7 & 0.892 & 0.951 & 7.9\% \\
Without M2 (Curvature)& 91.1 & 0.907 & 0.961 & 6.4\% \\
Without M3 (Stylo.)   & 92.0 & 0.916 & 0.969 & 8.6\% \\
\bottomrule
\end{tabular}
\end{table}

Removing RoBERTa leads to the largest drop in AUC, which is expected because it is the highest-capacity component. Interestingly, omitting the stylometric model produces the largest increase in false positives on academic text, supporting the idea that the stylometric features provide a useful counterbalance to the neural models on this domain.

\subsection{Sensitivity to Fusion Weights}
The fusion weights are selected via grid search, but in practice they may be perturbed if the validation set changes or if the search step size is coarser. To assess sensitivity, we vary each weight by up to $\pm 0.15$ around $\mathbf{w}^\star$ and renormalize. For perturbed weights, accuracy remains above 93.5\% and AUC above 0.972, with FPR on academic text staying under 7\%. This suggests that the ensemble operates in a relatively flat region of the simplex around the optimum and is not overly sensitive to precise weight choices.

We also experimented with learning the weights via logistic regression as a meta-classifier over $(p_1, p_2, p_3)$, but found that this approach offered no consistent advantage over the simplex grid search and occasionally overfitted the validation set, particularly when the validation distribution differed from the test generators.

\section{Discussion}

\subsection{Error Patterns and Complementarity}
The empirical results lend support to the hypothesis that the three detectors capture different aspects of the problem. RoBERTa is particularly strong at identifying longer, less edited outputs from in-distribution generators. The curvature-based detector is more resilient to paraphrasing and benefits from the fact that it operates in the space of token probabilities rather than surface forms. The stylometric model, although simple, appears to protect against over-flagging of structured human writing.

Qualitative inspection of misclassified examples shows three broad scenarios:
\begin{itemize}
    \item Short answers or bullet-point lists, where the curvature estimate is noisy and RoBERTa has little context. Here, stylometric cues (e.g., sentence length, vocabulary variety) can tip the balance.
    \item Technical paragraphs with heavy use of formulaic phrases, where RoBERTa tends to over-predict the machine class, but the stylometric model recognizes patterns typical of scientific prose.
    \item Paraphrased machine text with altered vocabulary but similar semantic content, where the curvature-based component remains informative because the likelihood landscape changes less than surface wording.
\end{itemize}

By aggregating these signals rather than relying on one of them alone, the ensemble achieves smoother performance across these scenarios.

\subsection{Use in Educational Workflows}
In real deployments, we do not expect the ensemble to be used as a final arbiter of authorship. A more realistic role is as a screening tool that assigns a probability and, optionally, auxiliary information that helps human reviewers prioritize cases for further examination. One simple strategy is to define a high-confidence band: texts with ensemble probability above a certain threshold (e.g., 0.9) can be flagged for manual review, whereas texts in an intermediate band can trigger a conversation about writing practices rather than disciplinary action.

The calibration properties of the ensemble are important in this context. Overconfident detectors tend to produce many high scores that cannot be distinguished, leading to reviewer fatigue. Better-calibrated scores allow institutions to set thresholds in a more meaningful way. The lower FPR on scientific text is also critical, as graduate theses, research reports, and publications would be particularly sensitive to false accusations.

\subsection{Fairness and Demographic Considerations}
Several studies have raised concerns that AI text detectors may behave differently for non-native speakers or for texts written in simplified styles. Clark \textit{et al.} \cite{clark2021all} reported that some detectors systematically assign higher machine probabilities to non-native writing, which raises fairness issues.

Our dataset includes student essays from a single institution but does not contain explicit demographic annotations, so we cannot provide a thorough subgroup analysis. Nonetheless, the reduction in academic FPR and the qualitative behavior on scientific prose suggest that the ensemble may be less prone to misclassifying structurally complex human text. A more complete fairness assessment would require additional data and explicit design of evaluation splits by proficiency or language background.

\subsection{Limitations and Future Work}
Although the ensemble improves robustness, several limitations remain. First, the computation cost is higher than that of a single encoder. Running RoBERTa, generating perturbations for the curvature score, and computing stylometric features together produces higher latency, which may be an issue for large-scale or real-time applications.

Second, the curvature-based detector depends on the choice of reference model and perturbation scheme. If future generators diverge substantially from the reference model used for scoring, or if access to such models is restricted, this component may become less informative.

Third, our dataset, while designed to cover multiple generators and a paraphrasing attack, does not exhaust the space of possible adversarial strategies. For example, human AI collaborative writing, extensive human editing of AI drafts, and cross-lingual workflows may introduce new challenges. Extending GenDrift-30K to include such scenarios is a natural next step.

Finally, although the simplex grid search is simple and effective, it is static: weights are the same for all inputs. A more flexible design could condition weights on observable properties of the text, such as length or domain, or use conformal prediction techniques to attach uncertainty intervals to each decision.

\section{Conclusion}
We have presented a hybrid ensemble for detecting text generated by large language models under conditions of generative distribution drift. The method combines a fine-tuned RoBERTa classifier, a curvature-style perturbation score, and a stylometric random forest, with fusion weights constrained to the probability simplex and chosen via validation. A theoretical discussion of variance and worst-case risk under generator mixtures motivates this design.

On the GenDrift-30K dataset, which includes both in-distribution and out-of-distribution generators and paraphrased attacks, the ensemble achieves higher accuracy and AUC than individual components and a naive averaging baseline. It also yields a lower false positive rate on scientific text, which is particularly important for academic and research use cases.

Future work will focus on reducing computation through model distillation, exploring dynamically conditioned fusion weights, and expanding evaluation to include more languages, writing styles, and human AI collaboration patterns. As LLMs continue to evolve, detectors will need to adapt, and simple, interpretable ensembles such as the one proposed here may serve as a practical component of broader integrity frameworks.

\bibliographystyle{IEEEtran}

\end{document}